\documentclass[10pt,letterpaper]{article}
\usepackage[margin=1.15in]{geometry}
\usepackage{times}
\usepackage[authoryear,round]{natbib}
\setcitestyle{citesep={;},aysep={,},yysep={;}}
\usepackage{xcolor}
\definecolor{lightgray}{gray}{0.93}
\usepackage[utf8]{inputenc}
\usepackage[T1]{fontenc}
\usepackage{amsmath,amssymb,graphicx,booktabs,multirow}
\usepackage{algorithm,algorithmic}
\usepackage{listings,tikz,colortbl,adjustbox,microtype}
\usepackage{environ,wrapfig,needspace}
\newsavebox{\tablewidthbox}
\NewEnviron{fitwidthtabular}[2]{%
  \sbox{\tablewidthbox}{\begin{tabular}{#2}\BODY\end{tabular}}%
  \ifdim\wd\tablewidthbox>#1\relax
    \resizebox{#1}{!}{\usebox{\tablewidthbox}}%
  \else
    \begin{tabular*}{#1}{@{\extracolsep{\fill}}#2}\BODY\end{tabular*}%
  \fi
}
\usepackage[hidelinks]{hyperref}
\usepackage{url}
\newcommand{\bcirc}[1]{%
  \tikz[baseline=(char.base)]{\node[shape=circle, fill=black, inner sep=1.2pt] (char)
    {\textcolor{white}{\scriptsize #1}};}}
\newcommand{\dreamicon}{\raisebox{-0.16em}{\includegraphics[height=1.15em]{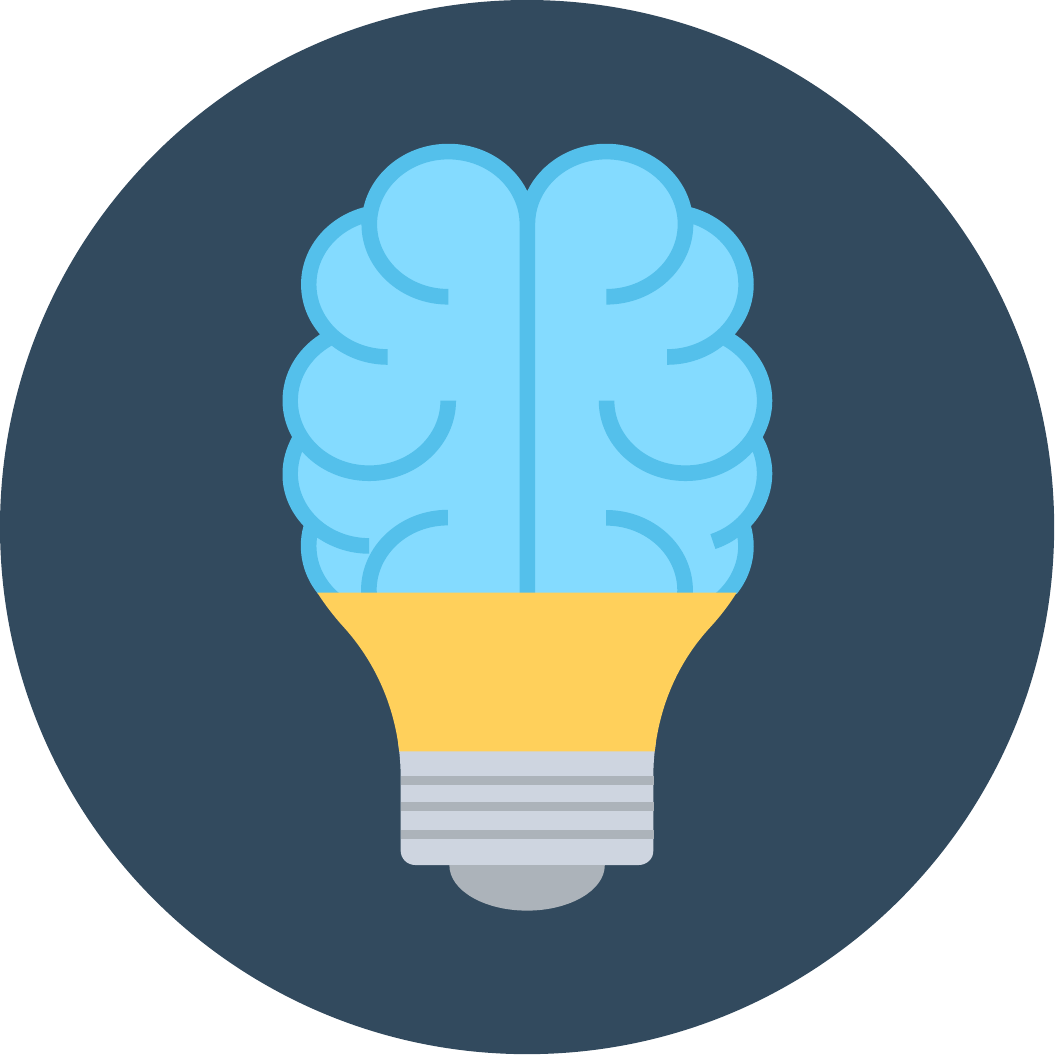}}}
\title{\dreamicon\enspace Remember Before You're Asked: MemDream for Self-Probing Memory Evolution}
\author{%
  Mingfei Lu \quad Mengjia Wu \quad Runsong Jia \quad Zhe Luo \quad Yi Zhang\\[0.5em]
  {\normalsize Australian Artificial Intelligence Institute (AAII)}\\
  {\normalsize University of Technology Sydney}\\[0.3em]
  {\small\texttt{\{mingfei.lu,runsong.jia,zhe.luo\}@student.uts.edu.au}}\\
  {\small\texttt{\{mengjia.wu,yi.zhang\}@uts.edu.au}}
}

\hypersetup{pdfauthor={Mingfei Lu, Mengjia Wu, Runsong Jia, Zhe Luo, Yi Zhang},
  pdftitle={Remember Before You're Asked: MemDream for Self-Probing Memory Evolution}}
\date{}
\begin{document}

\maketitle
\suppressfloats[t]

\begin{abstract}
Memory lets LLM-based agents stay coherent and personalized over long-horizon interactions. Stored facts can remain inaccessible when relevant evidence is fragmented, buried, or diluted during retrieval. Self-generated probing queries offer a direct signal for locating and repairing these latent failures. We ask whether a memory system can find and fix its own failures before they occur. We present \textsc{MemDream}, a framework for \emph{self-probing memory evolution}. During offline dream cycles, three specialized agents---Dreamer, Analyst, and Consolidator---probe, diagnose, and repair the memory graph before failures happen. A policy trained via Group Relative Policy Optimization learns which repair operations yield durable retrieval improvements, and a soft decay mechanism provides reversible forgetting driven by the same anticipatory signal. On LoCoMo and MemoryAgentBench (MAB), \textsc{MemDream} improves answer F1 by 3.98 points on LoCoMo and achieves a 9.1-point higher overall score on MAB over the strongest evaluated baselines. We hope self-probing memory evolution will serve as a solid baseline for agents that maintain their own memory.
\end{abstract}

\section{Introduction}
LLM-based agents now perform well in interactive settings such as multi-step reasoning, tool use, and open-ended dialogue~\citep{wang2024survey,li2024personal,schlegel2025large}. Over long-horizon interactions, an agent must accumulate, retain, and retrieve information across many turns as user goals and contexts shift~\citep{maharana2024evaluating}.  Memory carries this information forward, and it governs whether responses stay coherent and behavior stays consistent over time. But it can be wrong, and an agent is only as reliable as the memory it can correct before a failure reaches the user~\citep{zhang2025survey,pan2025secom}.

Prior work on long-horizon memory for LLM agents falls into three lines.
(1)~\emph{Structured memory architectures} organize information into
hierarchical tiers or graphs, which enriches retrieval but fixes the
organization in advance~\citep{packer2024memgpt,zhong2024memorybank,fluxmem2026,kang2025memoryos}.
(2)~\emph{Multi-stage retrieval pipelines} combine sparse, dense, and
reranking stages for precision, but leave the stored memory
unchanged~\citep{gutierrez2025rag,asai2024self}.
(3)~\emph{Self-evolving memory methods} reflect over accumulated
observations to synthesize higher-level memories~\citep{park2023generative,shinn2023reflexion},
self-question stored content to reorganize it~\citep{yang2026beyond},
tune retrieval configurations from observed failures~\citep{liu2026evolvemem},
or learn ADD / UPDATE / DELETE policies with reinforcement
learning~\citep{yan2025memory,wang2025memalpha,yuan2025memsearcher,zhang2026memrl,zhang2025memevolve,cai2025flex,zhai2025agentevolver}.
These advances motivate a complementary question: how can a memory system
construct a repair signal that identifies information it stores but cannot
reliably retrieve?

\Needspace{24\baselineskip}
\begin{wrapfigure}{r}{0.5\textwidth}
  \centering
  \includegraphics[width=0.5\textwidth]{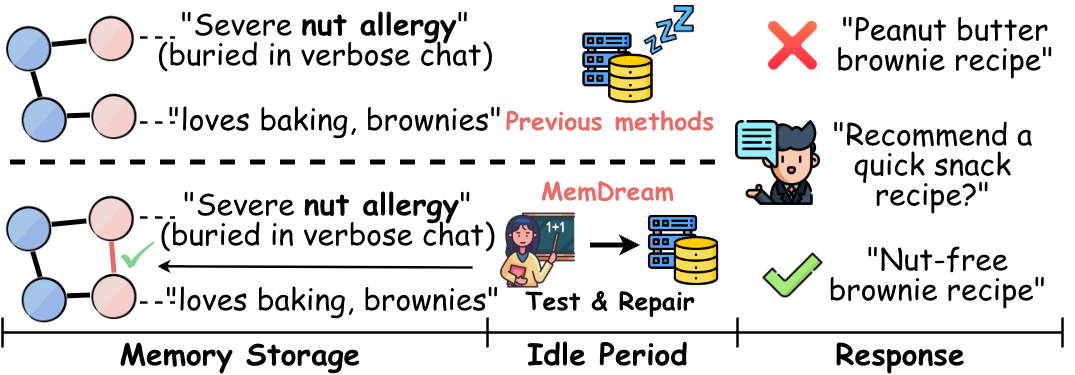}
  \vspace{-1em}
  \caption{(a) Existing reactive systems store a user's nut allergy and baking preferences as disconnected nodes, missing the dependency between them. When queried for a recipe, the system recommends a nut-containing dish, a harmful answer the user must catch. (b) Through asynchronous background consolidation, \textsc{MemDream}'s Dream Cycle finds and repairs this missing link, preventing the unsafe recommendation before the query arrives.}
  \label{fig:intro_comparison}
  \vspace{-1em}
\end{wrapfigure}

\textbf{By the time a real query exposes a weakness, the user has already
hit it.} As interactions accumulate, facts fragment across disconnected
nodes, links between sessions go missing, and entries grow stale.
Self-generated probing questions can expose these defects before a user
encounters them: the system tests whether relevant evidence is retrievable,
diagnoses the failure, and repairs the affected region. Reversible forgetting retains demoted content for later queries~\citep{wang2025selective}.

To address this, we present \textsc{MemDream}, a framework for
\emph{self-probing memory evolution}. The key insight: a memory system
should test itself. Inspired by memory consolidation during sleep~\citep{kumaran2016learning},
\textsc{MemDream} enters offline dream cycles in which three agents work
in turn: a Dreamer probes the memory graph with forward-looking queries,
an Analyst diagnoses where retrieval breaks down, and a Consolidator
repairs the graph before a real query can fail. A policy trained with
Group Relative Policy Optimization (GRPO)~\citep{shao2024deepseekmath,guo2025deepseek} replaces hand-crafted
heuristics, learning which repair operations yield durable retrieval
gains. A soft decay mechanism adds reversible forgetting, driven by the
same anticipatory signal. Together, these let \textsc{MemDream} fix weak
regions before a real query reaches them, rather than after. On LoCoMo
and MemoryAgentBench (MAB), it improves answer F1 by 3.98 points on LoCoMo
and overall score by 9.1 points on MAB over the strongest evaluated
baselines, FluxMem and O-Mem, respectively. Our
contributions are:

\noindent\bcirc{1} \textbf{Multi-Agent Dream Cycle.} Three agents (Dreamer, Analyst, Consolidator) generate forward-looking probes that turn unobserved future queries into a repair training signal.

\noindent\bcirc{2} \textbf{Learned Repair Policy.} GRPO trains the Consolidator on changes in retrieval success for the Dreamer's queries, replacing hand-tuned heuristics and assigning credit before user queries arrive.

\noindent\bcirc{3} \textbf{Reversible Selective Forgetting.} The same anticipatory signal drives soft decay: nodes are demoted rather than deleted and restored when later cycles reveal renewed relevance.

\begin{figure}[t]
  \centering
  \includegraphics[width=\textwidth]{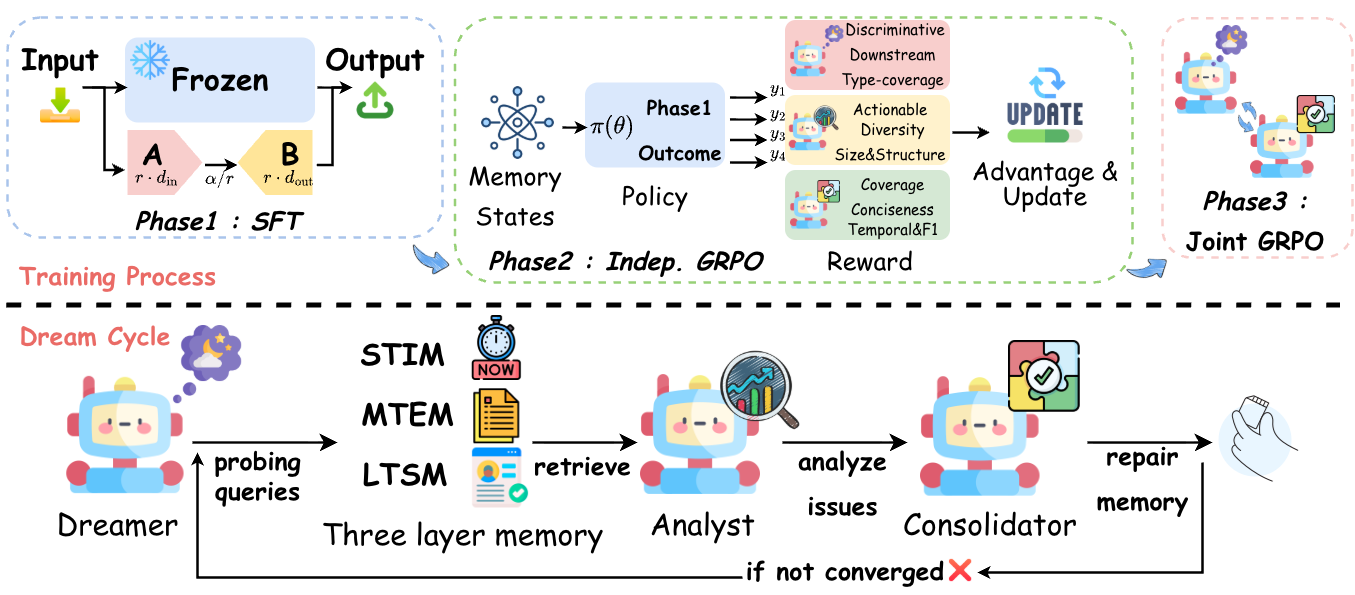}
  \caption{Overview of \textsc{MemDream}. Three LoRA-specialized agents share
  one Llama-3.1-8B backbone: the Dreamer generates probing queries aimed at
  weak regions, the Analyst diagnoses failures along four axes (buried,
  scattered, redundant, noisy), and the Consolidator applies repairs (extract,
  merge, synthesize, prune). The cycle repeats until retrieval F1 converges.
  We first train the agents independently with GRPO, then jointly fine-tune
the Dreamer and Consolidator (with the Analyst frozen) using end-to-end
memory quality as the reward.}
  \label{fig:dream_cycle}
\end{figure}

\section{Related Work}

\paragraph{Memory Architectures for LLM Agents.}
Memory designs for LLM agents have evolved from flat retrieval over vector
stores~\citep{lewis2020rag,gao2023retrieval} to hierarchical and structured
architectures~\citep{edge2024local,sarthi2024raptor}: MemGPT virtualizes context across tiers~\citep{packer2024memgpt};
MemoryBank adds Ebbinghaus-style strengthening and
decay~\citep{zhong2024memorybank}; MemoryOS adopts OS-inspired storage and
eviction~\citep{kang2025memoryos}; FluxMem selects memory structures by
interaction characteristics~\citep{fluxmem2026}; and MAGMA organizes memory
as multi-graphs for relational retrieval~\citep{jiang2026magma}. These
architectures shape \emph{how memory is organized and retrieved}, but leave
content maintenance to implicit or hand-crafted rules. \textsc{MemDream} is
complementary: it operates on top of such graphs, repairing their content
through anticipatory dream cycles rather than changing the storage layer.
At the context level, SEER combines visual compression with selective retrieval
of query-relevant text~\citep{xu2026seer}, addressing the cost of accessing
evidence within long inputs.

\paragraph{Self-Evolving and Policy-Based Memory.}
Adaptive memory systems learn update policies from task feedback~\citep{yan2025memory,wang2025memalpha,yuan2025memsearcher,zhang2026memrl},
reflect on stored experience~\citep{park2023generative,yang2026beyond}, or
consolidate memory between sessions~\citep{claude_dreaming2026}.
Recent methods target complementary objectives: Auto-Dreamer learns offline
region rewriting through GRPO~\citep{ye2026autodreamer}; TrustMem verifies
memory transitions for coverage, preservation, and
faithfulness~\citep{yang2026trustmem}; and RecMem reduces construction costs
through recurrence-triggered consolidation~\citep{dai2026recmem}.
\textsc{MemDream} focuses on the repair signal: self-generated queries
expose latent retrieval failures, whose diagnoses guide targeted repairs.
Appendix~\ref{app:consolidation_comparison} provides a detailed comparison.

\paragraph{Selective Forgetting.}
Continual learning has recognized the need to forget
selectively~\citep{kirkpatrick2017overcoming,parisi2019continual,shin2017continual}; in LLMs,
this appears as weight-level unlearning~\citep{jin2024machine,das2024larimar}, fine-grained
selective forgetting~\citep{wang2025selective}, and energy-based continual
learning~\citep{li2020energy}. Closest to our setting, FSFM combines passive
decay, active deletion, safety-triggered removal, and RL-based forgetting in
one hybrid pipeline~\citep{gu2026fsfm}. All of these treat forgetting as a
standalone operation with its own rules or scores. Our \emph{augment\_decay}
instead runs on the \emph{same} anticipatory signal as consolidation: a node
decays only when it is predicted to hurt future retrieval, and returns when a
later dream cycle finds it relevant. Forgetting is thus reversible within the dream cycle, rather than a separate, permanent step.

\paragraph{Structured Evidence and Diagnostic Selection.}
Related ideas also arise in task-specific reasoning and evidence selection.
JurisMA decomposes legal queries into element graphs and coordinates specialized
agents~\citep{lu2026query}; bias-aware citation prediction combines multi-agent
feature extraction with graph representations to assess papers without early
citation signals~\citep{lu2026newborn}. In graph fraud detection,
diffusion-guided learning augments features under sparse
supervision~\citep{liu2026beyond}, while PREF-Gate constrains label provenance
and uses validation to select relational evidence~\citep{liu2026pref}.
For LLM serving, Bottleneck-Preserving Witnessing selects compact workload
sets that expose component-level bottlenecks and verifies them through direct
measurements~\citep{liu2026diagnose}. These studies connect structured evidence,
selection, and verification across different tasks. MemDream applies these
concerns to persistent conversational memory, using prospective queries to
locate retrieval weaknesses and direct memory edits.

\section{Methodology}

\textsc{MemDream} tests its memory and repairs its weaknesses through
asynchronous background dream cycles (Figure~\ref{fig:dream_cycle}).
The agent continues answering queries using its current conversational
context and existing memory, while recent interactions remain available
in context during consolidation. Online interaction does not wait for a
dream cycle to complete.
It has two parts: a hierarchical memory store and a three-agent dream cycle
that probes and repairs it. We define the problem and memory architecture, then describe the dream cycle and GRPO-based policy learning.

\subsection{Problem Formalization}
\label{sec:formalization}

We consider a long-horizon setting where an agent converses with a user over
many turns. Let $\mathcal{H}_t = \{p_1, \ldots, p_t\}$ be the dialogue history
up to turn $t$, where each page $p_i = (u_i, a_i)$ is a user utterance paired
with an agent response. Memory is a hierarchical store
$\mathcal{G} = (\mathcal{V}, \mathcal{E})$ with three tiers: short-term
interaction memory (STIM) buffers recent pages with LRU eviction, mid-term
episodic memory (MTEM) groups pages into sessions with session-level
metadata, and long-term semantic memory (LTSM) holds compressed knowledge
entries. A node $v \in \mathcal{V}$ is a memory page, carrying a content
embedding, an entity set, access statistics, and a retrieval weight
$w_v \in [0, 1]$. An edge $e \in \mathcal{E}$ connects nodes that share
entities, belong to the same session, or were merged during consolidation.
At turn $t$, given a user query $q_t$, the agent retrieves relevant context
$\mathcal{C}_t = \text{Retrieve}(q_t, \mathcal{G})$ and generates a response
$y_t = \text{Generate}(q_t, \mathcal{C}_t)$. Retrieval uses reciprocal rank
fusion (RRF) over dense and sparse retrievers, weighted by each node's
retrieval weight:
\begin{equation}
s(v, q) = \left(\sum_{i \in \{\text{dense},\, \text{BM25}\}} \frac{1}{k + \text{rank}_i(v, q)}\right) \cdot w_v,
\label{eq:retrieval_score}
\end{equation}
where $\text{rank}_i(v, q)$ is the rank of node $v$ under retriever $i$,
$k$ is a smoothing constant, and $w_v \in [0,1]$ is the node's retrieval
weight (modulated by the dream cycle). Top-ranked nodes after cross-encoder
reranking form the context $\mathcal{C}_t$. The memory is then updated:
$\mathcal{G}_{t+1} = \text{Update}(\mathcal{G}_t, p_t)$.

Left to accumulate passively, $\mathcal{G}$ develops three failure modes:
\textbf{fragmentation} (related information scattered across disconnected
nodes), \textbf{redundancy} (overlapping content in multiple nodes), and
\textbf{staleness} (outdated information conflicting with recent interactions).
These compound over time, lowering retrieval precision.
We formalize memory evolution as learning a policy
$\pi: \mathcal{G} \mapsto \mathcal{G}'$ that maximizes expected retrieval
success over a distribution of likely future queries:
\begin{equation}
\mathbb{E}_{q \sim \mathcal{Q}}\left[\text{Success}(q,\, \text{Retrieve}(q,\, \mathcal{G}'))\right],
\label{eq:objective}
\end{equation}
where $\mathcal{Q}$ denotes a distribution of plausible future queries, not
a corpus of past observed failures. Since $\mathcal{Q}$ is unknown, the dream
cycle (\S\ref{sec:dream_cycle}) learns to draw samples from it, converting an
unobservable distribution into a tractable training signal.

\subsection{Multi-Agent Dream Cycle}
\label{sec:dream_cycle}

The dream cycle periodically probes the memory graph for weaknesses,
diagnoses their causes, and repairs them before real queries expose them.
Three agents divide this work over one LLM backbone, each with its own
LoRA adapter: the Dreamer probes, the Analyst diagnoses, and the Consolidator
repairs. The shared backbone keeps all three lightweight, and the separate
adapters let each learn a distinct role. Adapter and quantization settings
appear in the experimental setup.

\paragraph{Dreamer.} Generates probing queries that target weak regions of the
memory graph. The Dreamer first computes a weakness score for each node:
\begin{align}
\phi(v) = \;& \lambda_1 \Bigl(1 - \frac{\deg(v)}{\max_{u}\deg(u)}\Bigr) \notag \\
& + \lambda_2 \cdot \max_{u \in \mathcal{N}(v)} \cos(\mathbf{e}_v, \mathbf{e}_u) \notag \\
& + \lambda_3 \cdot \frac{t_{\text{now}} - t_v}{T},
\label{eq:weakness}
\end{align}
where the three terms capture connectivity deficit (isolated nodes),
redundancy (high similarity to a neighbor), and staleness (time since last
access), respectively. We select nodes with $\phi(v)$ above a threshold
as targets. Because $\phi(v)$ scores nodes individually, we then pair each
target $v^\star$ with a \emph{bridge partner}: the unlinked node most often co-retrieved with $v^\star$ across local probe queries $\mathcal{Q}(v^\star)$,
\begin{equation}
u^\star = \operatorname*{arg\,max}_{u:\,(u, v^\star) \notin \mathcal{E}}\;
\mathbb{E}_{q \sim \mathcal{Q}(v^\star)}\!\left[\mathbf{1}\!\left[\{u, v^\star\} \subseteq
\text{Retrieve}(q, \mathcal{G})\right]\right],
\label{eq:pair}
\end{equation}
where $\text{Retrieve}(q, \mathcal{G})$ is the retrieved set and
$\mathcal{E}$ the edge set. Pairing by co-retrievability rather than
embedding similarity links nodes that recur together yet lack an edge, and
the Dreamer then constructs a query that integrates $\{v^\star, u^\star\}$. For
instance, if separate nodes store a user's symptoms and medication history
with no connecting edge, the Dreamer generates a query requiring both. We pair
each probe with an \emph{expected answer} copied verbatim from a target
node's stored content rather than generated by the model, so the repair
signal stays grounded in existing memory and cannot chase a hallucinated target.
We train the Dreamer with SFT on (memory state, effective query) pairs, then
GRPO with retrieval success as reward.

\Needspace{15\baselineskip}
\begin{wraptable}{r}{0.5\textwidth}
\centering\footnotesize
\caption{Repair operation usage over the full evaluation.}
\label{tab:analyst_usage}
\setlength{\tabcolsep}{3pt}

\begin{fitwidthtabular}{\linewidth}{p{0.56\linewidth}lr}
\toprule
Category (2$\times$2 cell) & Repair & Count \\
\midrule
buried (intra-node, recall)     & extract    & 176 \\
scattered (inter-node, recall)  & synthesize & 301 \\
redundant (inter-node, precision) & merge    & 59 \\
noisy (intra-node, precision)   & prune      & 301 \\
\midrule
Total & & 837 \\
\bottomrule
\end{fitwidthtabular}

\end{wraptable}

\paragraph{Analyst.} Receives each dream query, its retrieval results, and the
expected answer, then classifies the failure along two axes: whether it lies
within a single node or across nodes, and whether it costs recall or precision.
The four cells are \textbf{buried} (intra-node recall: a fact is present but
hidden in verbose text), \textbf{scattered} (inter-node recall: evidence split
across unlinked nodes), \textbf{redundant} (inter-node precision: overlapping
content across nodes), and \textbf{noisy} (intra-node precision: low-value nodes
such as greetings or filler). Each cell maps to one repair the
Consolidator can execute: extract, synthesize, merge, or prune. For each failure the Analyst emits the target node ids and the prescribed
operation, ordering them worst-first so the most damaging defects are repaired
first. Because the axes enumerate how retrieval can fail rather than list
hand-picked symptoms, the taxonomy is complete over these two axes and extends
by refining a cell into subcases without changing the Dreamer, the reward, or
the training pipeline. Across the evaluation all four cells are exercised
(Table~\ref{tab:analyst_usage}), so diagnosis does not collapse onto one category.

\paragraph{Consolidator.} Executes the repairs the Analyst prescribes. Extract
turns a verbose page into structured nodes. Synthesize generates summary nodes
with provenance links. Merge combines similar nodes, preserving edge
connectivity. Prune applies a soft decay rather than hard deletion:
\begin{equation}
w_v \leftarrow \gamma \cdot w_v, \quad \gamma = 0.2,
\label{eq:decay}
\end{equation}
where $\gamma$ is a retention factor that keeps a $\gamma$ fraction of the
node's weight: $\gamma{=}0$ deletes the node outright (the w/o Decay
ablation), while $\gamma$ near $1$ leaves it almost unchanged. Decay lowers the
node's retrieval ranking but keeps it accessible (Eq.~\ref{eq:retrieval_score}),
so pruning never loses information.
If a later cycle shows renewed relevance, the weight is restored:
\begin{equation}
w_v \leftarrow \min(1.0,\; w_v + \eta \cdot \text{success\_count}_v).
\label{eq:restore}
\end{equation}
Each dream round processes a different partition of memory nodes until convergence:
\begin{equation}
\frac{1}{|\mathcal{D}|}\sum_{q \in \mathcal{D}} \text{F1}(q, \mathcal{G}) \geq \tau \quad \text{or}
\quad t \geq T_{\max},
\label{eq:early_stop}
\end{equation}
where $\mathcal{D}$ denotes dream queries, $\tau$ the convergence threshold, and $T_{\max}$ the round limit.

\subsection{GRPO-Based Agent Training}
\label{sec:grpo}

We train the three dream-cycle agents via Group Relative Policy
Optimization (GRPO). Each agent optimizes its own reward: the Dreamer
for generating discriminative probes, the Analyst for accurate diagnosis,
and the Consolidator for repairs that improve retrieval. The agents share
the Llama-3.1-8B backbone through separate LoRA adapters.

\subsubsection{Per-Agent Rewards}

Each agent optimizes its own fixed-weight reward, a sum of interpretable terms:
\begin{align}
r_{\text{D}} &= w_{\text{disc}}\, r_{\text{disc}} + w_{\text{down}}\, r_{\text{down}}
    + w_{\text{cov}}\, r_{\text{cov}}, \label{eq:reward_dreamer}\\
r_{\text{A}} &= w_{\text{act}}\, r_{\text{act}} + w_{\text{div}}\, r_{\text{div}}
    + w_{\text{size}}\, r_{\text{size}} + w_{\text{struct}}\, r_{\text{struct}},
\label{eq:reward_analyst}\\
r_{\text{C}} &= w_{\text{qa}}\, r_{\text{qa}} + w_{\text{cov}}\, r_{\text{cov}}
    + w_{\text{temp}}\, r_{\text{temp}} + w_{\text{concise}}\, r_{\text{concise}}.
\label{eq:reward_consol}
\end{align}
The \textbf{Dreamer} terms favor probes that are discriminative, expose
downstream failures, and cover diverse query types: $r_{\text{disc}}$
rewards queries that separate weak from strong memory regions,
$r_{\text{down}}$ measures downstream QA impact, and $r_{\text{cov}}$
measures coverage over query types.
The \textbf{Analyst} terms favor diagnoses that are actionable, varied in
repair type, appropriately sized, and well-structured: $r_{\text{act}}$ is
the fraction of repairs that reference valid target nodes, $r_{\text{div}}$
rewards using distinct repair types, $r_{\text{size}}$ peaks at a moderate
number of repairs per cycle, and $r_{\text{struct}}$ rewards well-formed
output.
The \textbf{Consolidator} reward is operation-specific. For an extract
operation ($r_{\text{C}}$), $r_{\text{qa}}$ is QA-F1 on affected queries,
$r_{\text{cov}}$ measures retained coverage, $r_{\text{temp}}$ rewards
preserved temporal cues, and $r_{\text{concise}}$ penalizes bloat. Merge
and synthesize use analogous rewards that trade QA-F1 and coverage against
deduplication and grounding.
All weights are fixed constants, set a priori from each term's role
(primary objective versus regularizer) rather than tuned on the evaluation
benchmarks. Their values are given in the appendix.
For each training instance, we sample $N$ rollouts from the current
policy and normalize their advantages within the group:

\begin{align}
\mathcal{L}_{\text{GRPO}} &= -\frac{1}{N}\sum_{i=1}^{N}
A_i \cdot \log \pi_\theta(a_i \mid s), \notag \\
A_i &= \frac{r_i - \bar{r}}{\sigma_r + \epsilon},
\label{eq:grpo}
\end{align}

where $\bar{r}$ and $\sigma_r$ are the group mean and standard deviation
of rewards, and $\pi_\theta$ is the LoRA-adapted agent policy. Only
rollouts above $\bar{r}$ receive positive gradient. This yields a harder curriculum.

\subsubsection{Training Procedure}

Training proceeds in three phases on MSC dialogue data with
pre-generated dream-augmented queries. First, the Dreamer, Analyst, and
Consolidator are pre-trained via SFT on teacher data (GPT-4.1 outputs). Second, each
agent undergoes independent GRPO with its own reward
(Eqs.~\ref{eq:reward_dreamer}--\ref{eq:reward_consol}). Third, Dreamer
and Consolidator are jointly fine-tuned while Analyst remains frozen.
The joint reward measures memory improvement:
\begin{equation}
r_{\text{joint}} = \frac{1}{|\mathcal{D}_{\text{new}}|} \sum_{q \in \mathcal{D}_{\text{new}}}
\bigl[\text{F1}(q,\, \mathcal{G}') - \text{F1}(q,\, \mathcal{G})\bigr],
\label{eq:joint_reward}
\end{equation}
where $\mathcal{D}_{\text{new}}$ is a fresh set of Dreamer-generated queries
(not seen during independent training), $\mathcal{G}$ is the memory before
consolidation, and $\mathcal{G}'$ is the memory after. This rewards the
Dreamer for queries that expose fixable weaknesses, and the
Consolidator for repairs that durably improve retrieval. This joint reward
replaces the per-agent reward in Eq.~\ref{eq:grpo}, with gradients flowing only to the Dreamer and
Consolidator adapters.
The agents are trained with learning rate $1 \times 10^{-3}$,
group size $N{=}4$, and exploration rate $\epsilon{=}0.2$ that decays
over training. Full hyperparameters are provided in
Appendix~\ref{app:training}. These training stages are completed before
deployment. During deployment, model parameters remain fixed, and dream
cycles update only the external memory.

\begin{figure}[t]
  \centering
  \setlength{\abovecaptionskip}{4pt}
  \includegraphics[width=\textwidth,trim={0 20bp 0 0},clip]{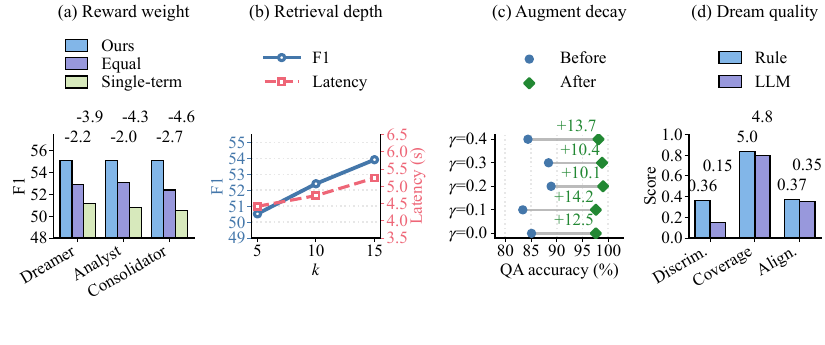}
  \caption{Sensitivity analysis. (a) Per-agent GRPO weights outperform equal weighting and single-term ablations. (b) Increasing retrieval depth $k$ improves F1 with diminishing returns and linear latency growth. (c) Dream cycles improve accuracy at every decay factor $\gamma$. (d) Rule-based probes lead on discrimination and alignment, with similar coverage.}
  \label{fig:analysis}
\end{figure}

\section{Experiments}

\subsection{Experimental Setup}
\label{sec:experimental_setup}

\paragraph{Datasets.}
We evaluate on two benchmarks.
\textbf{LoCoMo}~\citep{maharana2024evaluating} contains 10 long-horizon conversations and 1{,}986 human-written QA pairs (about 199 per conversation). Following the standard protocol, we evaluate 1{,}540 multi-hop, temporal, open-ended, and single-hop questions, excluding the adversarial category.
\textbf{MemoryAgentBench (MAB)}~\citep{hu2026evaluatingmemoryllmagents} covers four memory competencies:
Accurate Retrieval (AR), Test-Time Learning (TTL), Long-Range Understanding
(LRU), and Selective Forgetting (SF), across 9 sub-tasks including RULER QA,
EventQA, LongMemEval, ICL classification, Detective QA, InfBench
summarization, and FactConsolidation.

\paragraph{Baselines.}
We compare against ten memory systems:
\textbf{LangMem}~\citep{langmem2025}: a flat memory with embedding-based
retrieval.
\textbf{Mem0}~\citep{chhikara2025mem0}: a flat retrieval-based system with
incremental extraction and deduplication.
\textbf{ZEP}~\citep{rasmussen2025zep}: a time-aware knowledge graph memory
with entity-level timestamping.
\textbf{MemoryOS}~\citep{kang2025memoryos}: an OS-inspired memory with
tiered storage and eviction policies.
\textbf{A-Mem}~\citep{xu2025mem}: a structured memory using evolving linked
notes with self-reflection.
\textbf{O-Mem}~\citep{wang2025mem}: a graph-based memory with explicit
temporal ordering and fact-update mechanisms.
\textbf{MEMOS}~\citep{li2025memos}: a multi-granularity memory with
redundancy filtering.
\textbf{MemR$^3$}~\citep{du2025memr3}: a closed-loop controller alternating retrieval,
reflection, and reranking for evidence coverage.
\textbf{HippoRAG~2}~\citep{gutierrez2025rag}: a hippocampus-inspired knowledge-graph retrieval framework.
\textbf{FluxMem}~\citep{fluxmem2026}: an adaptive multi-structure memory
that learns to select among memory organizations per interaction.
All systems use the same reader LLM (GPT-4.1, temperature 0).

\paragraph{Implementation Details.}
MemDream uses a three-layer memory hierarchy (STIM capacity 4, MTEM up to
2000 sessions, LTSM up to 100 entries). The dream cycle deploys three
LoRA-adapted agents on Llama-3.1-8B-Instruct with 8-bit quantization (rank $r{=}16$, $\alpha{=}32$, applied to the attention projections; $\sim$39.3M trainable parameters total). All training and inference run on a single A100 (80GB).
Retrieval fuses BM25 and dense encoding (BAAI/bge-m3) via reciprocal rank
fusion, followed by cross-encoder reranking (bge-reranker-v2-m3). The agents
are trained via GRPO with group size $N{=}4$ on MSC dialogue data. On
LoCoMo, we use chunk size 4096 with the full dream consolidation pipeline.
On MAB, we use chunk size 512 for AR tasks (GPT-4.1 reader) and SF tasks,
chunk size 4096 for TTL, LRU, and AR tasks with the Llama-8B reader.

\begin{table}[t]
  \centering
  \footnotesize
  \setlength{\tabcolsep}{1.2pt}
  \renewcommand{\arraystretch}{1.15}
  \definecolor{fluxrow}{gray}{0.94}
  
\begin{fitwidthtabular}{\linewidth}{@{}l@{\hspace{4pt}}ccc@{\hspace{5pt}}ccc@{\hspace{5pt}}ccc@{\hspace{5pt}}ccc@{\hspace{8pt}}ccc@{}}
\toprule
Method & \multicolumn{3}{c}{Cat1: Multi-hop} & \multicolumn{3}{c}{Cat2: Temporal} & \multicolumn{3}{c}{Cat3: Open} & \multicolumn{3}{c}{Cat4: Single-hop} & \multicolumn{3}{c}{\textbf{Average}} \\
\cmidrule(lr{1pt}){2-4}\cmidrule(lr{1pt}){5-7}\cmidrule(lr{1pt}){8-10}\cmidrule(lr{1pt}){11-13}\cmidrule(lr{1pt}){14-16}
 & F1 & B-1 & R-L & F1 & B-1 & R-L & F1 & B-1 & R-L & F1 & B-1 & R-L & F1 & B-1 & R-L \\
\midrule
LangMem & 41.11 & 32.09 & 37.71 & 53.67 & 46.22 & 47.37 & 33.38 & 27.26 & 24.55 & 51.13 & 44.22 & 48.37 & 44.82 & 37.45 & 39.50 \\
Mem0 & 28.84 & 20.22 & 27.45 & 24.21 & 18.75 & 23.23 & 23.59 & 19.09 & 23.13 & 32.74 & 27.58 & 33.02 & 27.35 & 21.41 & 26.71 \\
ZEP & 8.74 & 5.98 & 6.79 & 7.40 & 4.70 & 6.07 & 7.64 & 3.46 & 5.18 & 7.91 & 3.80 & 6.26 & 7.92 & 4.49 & 6.08 \\
MemoryOS & 43.03 & 34.09 & 39.75 & 44.43 & 32.03 & 42.64 & 28.36 & 23.24 & 27.83 & 49.31 & 42.61 & 49.84 & 41.28 & 32.99 & 40.02 \\
A-Mem & 32.41 & 24.28 & 31.26 & 37.23 & 33.02 & 39.61 & 16.16 & 14.64 & 15.19 & 40.94 & 35.54 & 42.32 & 31.69 & 26.87 & 32.10 \\
MEMOS & 20.27 & 13.35 & 15.75 & 14.86 & 7.86 & 11.58 & 11.87 & 6.85 & 9.32 & 18.96 & 11.42 & 17.37 & 16.49 & 9.87 & 13.51 \\
MemR$^{3}$ & 18.65 & 11.17 & 14.37 & 25.39 & 15.90 & 22.63 & 11.56 & 5.77 & 8.96 & 24.35 & 12.72 & 22.00 & 19.99 & 11.39 & 16.99 \\
O-Mem & 42.64 & 34.08 & 32.78 & \underline{57.48} & \textbf{49.76} & 48.88 & 30.58 & 25.69 & 21.11 & 54.89 & 48.98 & 44.26 & 46.40 & 39.63 & 36.76 \\
HippoRAG 2 & 28.11 & 18.92 & 27.68 & 10.35 & 6.92 & 9.82 & 24.91 & 21.10 & 24.48 & 44.82 & 37.25 & 44.37 & 27.05 & 21.05 & 26.59 \\
\rowcolor{fluxrow}
FluxMem & \underline{48.56} & \underline{39.90} & \underline{44.05} & 56.67 & 41.88 & \underline{54.89} & \textbf{37.30} & \underline{29.63} & \textbf{36.76} & \underline{62.12} & \underline{53.52} & \underline{62.33} & \underline{51.16} & \underline{41.23} & \underline{49.51} \\
\textbf{MemDream} & \textbf{53.64} & \textbf{43.88} & \textbf{47.50} & \textbf{66.05} & \underline{49.07} & \textbf{63.43} & \underline{35.89} & \textbf{30.69} & \underline{35.43} & \textbf{64.97} & \textbf{57.66} & \textbf{65.12} & \textbf{55.14} & \textbf{45.33} & \textbf{52.87} \\
\midrule
    \rowcolor{lightgray}
    \textbf{Improv.} & +5.08 & +3.98 & +3.45 & +8.57 & -0.69 & +8.54 & -1.41 & +1.06 & -1.33 & +2.85 & +4.14 & +2.79 & +3.98 & +4.10 & +3.36 \\
\bottomrule
\end{fitwidthtabular}

\caption{LoCoMo results (GPT-4.1 reader). Best in bold; second-best underlined.}
  \label{tab:locomo_main}
\end{table}

\begin{table}[t]
  \centering
  \footnotesize
  \setlength{\tabcolsep}{1.2pt}
  \renewcommand{\arraystretch}{1.15}
  \definecolor{fluxrow}{gray}{0.94}
  
\begin{fitwidthtabular}{\linewidth}{@{}l@{\hspace{4pt}}ccc@{\hspace{5pt}}ccc@{\hspace{5pt}}ccc@{\hspace{5pt}}ccc@{\hspace{8pt}}ccc@{}}
\toprule
Method & \multicolumn{3}{c}{Cat1: Multi-hop} & \multicolumn{3}{c}{Cat2: Temporal} & \multicolumn{3}{c}{Cat3: Open} & \multicolumn{3}{c}{Cat4: Single-hop} & \multicolumn{3}{c}{\textbf{Average}} \\
\cmidrule(lr{1pt}){2-4}\cmidrule(lr{1pt}){5-7}\cmidrule(lr{1pt}){8-10}\cmidrule(lr{1pt}){11-13}\cmidrule(lr{1pt}){14-16}
 & F1 & B-1 & R-L & F1 & B-1 & R-L & F1 & B-1 & R-L & F1 & B-1 & R-L & F1 & B-1 & R-L \\
\midrule
Mem0 & 9.96 & 6.25 & 7.69 & 10.73 & 5.99 & 8.86 & 7.23 & 5.08 & 6.00 & 18.40 & 11.21 & 16.76 & 11.58 & 7.13 & 9.83 \\
ZEP & 15.04 & 11.56 & 2.89 & 3.49 & 2.68 & 2.78 & \textbf{26.67} & 18.44 & 2.57 & 30.15 & 17.15 & 3.37 & 18.84 & 12.46 & 2.90 \\
MemoryOS & 13.60 & 10.57 & 24.91 & 22.18 & 18.68 & 29.85 & 9.11 & 8.21 & 17.60 & 24.81 & 21.09 & 38.14 & 17.43 & 14.64 & 27.63 \\
A-Mem & 20.46 & 14.37 & 21.53 & 25.33 & 22.01 & 28.45 & 15.62 & 13.72 & 14.95 & 33.45 & 28.85 & 37.07 & 23.72 & 19.74 & 25.50 \\
MEMOS & 33.00 & 24.19 & 32.40 & 42.16 & 34.66 & 43.26 & 17.97 & 12.79 & 16.15 & 41.91 & 33.01 & \underline{48.89} & 33.76 & 26.16 & 35.18 \\
MemR$^{3}$ & 9.27 & 4.52 & 5.92 & 7.47 & 2.76 & 4.48 & 6.50 & 2.72 & 3.73 & 12.07 & 4.41 & 8.33 & 8.83 & 3.60 & 5.62 \\
O-Mem & \underline{35.62} & 27.27 & \underline{32.78} & \underline{49.77} & \underline{37.21} & \underline{48.88} & 21.41 & 15.97 & \underline{21.11} & \underline{44.08} & \underline{38.26} & 44.26 & \underline{37.72} & 29.68 & \underline{36.76} \\
HippoRAG 2 & 23.23 & \textbf{29.48} & 21.47 & 7.55 & 9.77 & 7.46 & 19.77 & \textbf{27.67} & 19.30 & 34.85 & 35.79 & 34.35 & 21.35 & 25.68 & 20.65 \\
\rowcolor{fluxrow}
FluxMem & 25.61 & \underline{29.22} & 23.85 & 30.54 & 32.17 & 29.93 & 18.26 & 24.03 & 17.39 & 37.58 & 38.19 & 36.43 & 28.00 & \underline{30.90} & 26.90 \\
\textbf{MemDream} & \textbf{39.00} & 25.09 & \textbf{34.85} & \textbf{52.70} & \textbf{37.57} & \textbf{50.64} & \underline{22.48} & \underline{25.50} & \textbf{22.17} & \textbf{52.02} & \textbf{42.79} & \textbf{52.34} & \textbf{41.55} & \textbf{32.74} & \textbf{40.00} \\
\midrule
    \rowcolor{lightgray}
    \textbf{Improv.} & +3.38 & -4.39 & +2.07 & +2.93 & +0.36 & +1.76 & -4.19 & -2.17 & +1.06 & +7.94 & +4.53 & +3.45 & +3.83 & +1.84 & +3.24 \\
\bottomrule
\end{fitwidthtabular}

    \caption{LoCoMo results (Llama-3.1-8B reader). Best in bold; second-best underlined.}
  \label{tab:locomo_llama}
\end{table}

\paragraph{Evaluation Metrics.}
On LoCoMo, we report F1, BLEU-1 (B-1), and ROUGE-L (R-L) per category;
Average equally weights the four categories. On
MAB, we follow the original protocol: substring accuracy for AR, SF, TTL,
and Detective QA, and ROUGE-L for summarization. We report results with
deterministic decoding (temperature 0), following the LoCoMo and MAB
protocols. The gains are stable under bootstrap confidence
intervals and repeated runs; full per-category intervals are given in the
appendix.

\subsection{Overall Performance}

Tables~\ref{tab:locomo_main},~\ref{tab:locomo_llama},~\ref{tab:mab_gpt41}, and~\ref{tab:mab_llama}
report results on LoCoMo and MAB.

\paragraph{LoCoMo.}
With both GPT-4.1 and Llama readers (Tables~\ref{tab:locomo_main}
and~\ref{tab:locomo_llama}), MemDream reaches the highest average F1, B-1,
and R-L. With GPT-4.1, it gains 3.98 average F1 over FluxMem and 8.57 temporal F1
over O-Mem, the strongest baselines for the respective metrics. It trails FluxMem on open-ended questions by 1.41 F1;
these favor broad context over targeted probes.

\paragraph{MAB.}
Across both readers (Tables~\ref{tab:mab_gpt41}
and~\ref{tab:mab_llama}), MemDream reaches the highest overall score, 54.2
with GPT-4.1 and 37.1 with Llama, and leads on 7 of 9 tasks with GPT-4.1.
It improves on Accurate Retrieval (SH-QA 67.0, MH-QA 56.0, LME 28.7,
EventQA 81.6) and Test-Time Learning (91.0), where the dream cycle links
memories across sessions. Two tasks favor other systems: MemoryOS leads
Detective QA (73.2 vs 71.8), where hierarchical indexing fits long
narratives, and O-Mem leads Summarization (18.2 vs 17.4), where the dream
cycle trades verbatim coverage for factual precision and loses ROUGE
overlap. With the Llama reader, MemDream leads the second-best
(FluxMem 25.6) by 11.5, with the largest gains on Accurate
Retrieval and TTL.

\begin{table}[t]
  \centering
  \small
  \setlength{\tabcolsep}{3.5pt}
  
\begin{fitwidthtabular}{\linewidth}{l|cccc|c|cc|cc|c}
    \toprule
    \multirow{2}{*}{Method}
    & \multicolumn{4}{c|}{AR}
    & TTL
    & \multicolumn{2}{c|}{LRU}
    & \multicolumn{2}{c|}{SF}
    & \multirow{2}{*}{Overall} \\
    & SH-QA & MH-QA & LME & EventQA
    & MCC
    & Summ. & Det QA
    & FC-SH & FC-MH & \\
    \midrule
    A-Mem & 24.0 & 38.0 & 6.0 & 68.0 & 29.0 & 15.6 & 52.1 & 8.0 & 2.0 & 27.0 \\
    HippoRAG 2 & 30.0 & 46.0 & 7.7 & 69.6 & 54.0 & 16.2 & 47.9 & 18.0 & 6.0 & 32.8 \\
    MemoryOS & 48.0 & 45.0 & 9.0 & 70.6 & 27.0 & 16.9 & \textbf{73.2} & 11.0 & 5.0 & 34.0 \\
    MEMOS & 32.0 & \underline{49.0} & 6.0 & 66.2 & 55.0 & 16.1 & 49.3 & 8.0 & 1.0 & 31.4 \\
    MemR$^{3}$ & 27.0 & 40.0 & 6.7 & 66.6 & 47.0 & 16.1 & 49.3 & 16.0 & 4.0 & 30.3 \\
    O-Mem & \underline{62.0} & 2.0 & \underline{26.7} & \underline{76.0} & \underline{85.0} & \textbf{18.2} & 69.0 & \underline{56.0} & \underline{11.0} & \underline{45.1} \\
    \midrule
    \rowcolor{lightgray}
    \textbf{MemDream} & \textbf{67.0} & \textbf{56.0} & \textbf{28.7} & \textbf{81.6} & \textbf{91.0} & \underline{17.4} & \underline{71.8} & \textbf{61.0} & \textbf{13.3} & \textbf{54.2} \\
    \bottomrule
  \end{fitwidthtabular}

  \caption{Performance on MAB (GPT-4.1 reader). Metrics follow the MAB protocol. Best in bold, second-best underlined. We compare only memory agent methods under identical settings.}
  \label{tab:mab_gpt41}
\end{table}

\begin{table}[t]
  \centering
  \small
  \setlength{\tabcolsep}{3.5pt}
  
\begin{fitwidthtabular}{\linewidth}{l|cccc|c|cc|cc|c}
    \toprule
    \multirow{2}{*}{Method}
    & \multicolumn{4}{c|}{AR}
    & TTL
    & \multicolumn{2}{c|}{LRU}
    & \multicolumn{2}{c|}{SF}
    & \multirow{2}{*}{Overall} \\
    & SH-QA & MH-QA & LME & EventQA
    & MCC
    & Summ. & Det QA
    & FC-SH & FC-MH & \\
    \midrule
    FluxMem & 19.0 & \underline{26.0} & 6.7 & 33.3 & \underline{80.0} & 15.5 & 16.7 & 28.0 & 5.0 & \underline{25.6} \\
    HippoRAG 2 & 10.0 & 21.0 & 6.0 & 31.8 & 40.0 & 14.4 & 53.5 & 22.0 & \textbf{7.0} & 22.9 \\
    MemoryOS & 19.0 & 8.0 & 4.3 & 31.0 & 13.0 & 8.2 & 47.9 & 2.0 & 1.0 & 14.9 \\
    MEMOS & 3.0 & 24.0 & 5.0 & 32.0 & 45.0 & 25.5 & \textbf{59.2} & 24.0 & 1.0 & 24.3 \\
    MemR$^{3}$ & 10.0 & 22.0 & 5.7 & \underline{38.0} & 43.0 & \textbf{27.1} & 40.8 & 17.0 & 3.0 & 23.0 \\
    O-Mem & \underline{20.6} & 22.0 & \underline{13.3} & 27.3 & 36.0 & 11.4 & 33.3 & \textbf{45.0} & \underline{6.0} & 23.9 \\
    \midrule
    \rowcolor{lightgray}
    \textbf{MemDream} & \textbf{43.0} & \textbf{35.0} & \textbf{20.0} & \textbf{39.0} & \textbf{84.0} & \underline{26.3} & \underline{56.0} & \underline{29.0} & 2.0 & \textbf{37.1} \\
    \bottomrule
  \end{fitwidthtabular}

  \caption{MAB results (Llama-3.1-8B; metrics as in Table~\ref{tab:mab_gpt41}). Best in bold; second-best underlined.}
  \label{tab:mab_llama}
\end{table}

\begin{table}[!t]
  \centering
  \footnotesize
  \setlength{\tabcolsep}{1.2pt}
  \renewcommand{\arraystretch}{1.15}
  
\begin{fitwidthtabular}{\linewidth}{@{}l@{\hspace{4pt}}ccc@{\hspace{5pt}}ccc@{\hspace{5pt}}ccc@{\hspace{5pt}}ccc@{\hspace{8pt}}ccc@{}}
\toprule
Variant & \multicolumn{3}{c}{Cat1: Multi-hop} & \multicolumn{3}{c}{Cat2: Temporal} & \multicolumn{3}{c}{Cat3: Open} & \multicolumn{3}{c}{Cat4: Single-hop} & \multicolumn{3}{c}{\textbf{Average}} \\
\cmidrule(lr{1pt}){2-4}\cmidrule(lr{1pt}){5-7}\cmidrule(lr{1pt}){8-10}\cmidrule(lr{1pt}){11-13}\cmidrule(lr{1pt}){14-16}
 & F1 & B-1 & R-L & F1 & B-1 & R-L & F1 & B-1 & R-L & F1 & B-1 & R-L & F1 & B-1 & R-L \\
\midrule
\textbf{MemDream (full)} & \textbf{53.64} & \textbf{43.88} & \textbf{47.50} & \textbf{66.05} & \textbf{49.07} & \textbf{63.43} & \textbf{35.89} & \textbf{30.69} & \textbf{35.43} & \textbf{64.97} & \textbf{57.66} & \textbf{65.12} & \textbf{55.14} & \textbf{45.33} & \textbf{52.87} \\
\quad w/o Dreamer & 49.81 & 40.22 & 44.17 & 62.43 & 46.51 & 59.87 & 34.72 & 28.93 & 33.61 & 61.38 & 54.12 & 61.29 & 52.09 & 42.45 & 49.74 \\
\quad w/o Analyst & 51.47 & 42.15 & 45.82 & 63.89 & 47.64 & 61.05 & 35.11 & 29.48 & 34.26 & 63.25 & 55.71 & 63.48 & 53.43 & 43.75 & 51.15 \\
\quad w/o Consolidator GRPO & 50.93 & 41.67 & 45.21 & 63.18 & 47.12 & 60.54 & 34.95 & 29.17 & 33.89 & 62.84 & 55.28 & 62.91 & 52.98 & 43.31 & 50.64 \\
\quad w/o Decay & 39.44 & 27.07 & 35.82 & 52.96 & 38.80 & 51.22 & 17.35 & 12.51 & 17.89 & 54.41 & 44.67 & 54.70 & 41.04 & 30.76 & 39.91 \\
\bottomrule
\end{fitwidthtabular}

  \caption{Ablation study. Best results in each column are shown in bold.}
  \label{tab:ablation}
\end{table}

\subsection{Ablation Study}
\label{sec:ablations}

We test four dream-cycle ablations on LoCoMo (GPT-4.1 reader), removing or degrading one component at a time:
(1)~\textbf{w/o Dreamer}: no probing queries; the system keeps the initial
memory state without dream-guided maintenance.
(2)~\textbf{w/o Analyst}: the Consolidator gets no targeted diagnosis and
extracts uniformly from all nodes above a length threshold.
(3)~\textbf{w/o Consolidator GRPO}: the Consolidator keeps its
SFT-initialized weights, without GRPO fine-tuning.
(4)~\textbf{w/o Decay} ($\gamma{=}0$): prune permanently removes nodes
instead of applying soft decay.
Table~\ref{tab:ablation} reports the results.

\paragraph{Results.}
The largest effect comes from augment\_decay. Removing it (w/o Decay) drops
average F1 from 55.14 to 41.04 ($-$14.10), and by 18.54 F1 on open-ended
questions. Hard pruning discards nodes that later sessions need, so
reversible forgetting is what keeps long-horizon memory intact.

Among the three agents, the Dreamer matters most: removing it drops F1 by
3.05, leaving the consolidation cycle with no self-generated signal to
target. Removing the Analyst drops F1 by 1.71, as the Consolidator then
extracts uniformly, without diagnosis. Removing GRPO drops F1 by 2.16; the
SFT-only Consolidator still repairs, but without a learned policy for
choosing operations. The full system leads on every category and metric,
so the three agents provide complementary value.

\paragraph{Offline computational cost.}
The approximate timing profile in Appendix~\ref{app:cost} assigns 2.1, 1.5,
and 3.2 minutes per round to the Dreamer, Analyst, and Consolidator,
respectively, with another 1.4 minutes for retrieval and verification.
A round takes about 9 minutes including model and adapter loading;
a full cycle is reported at about 22 minutes for 2--3 rounds.
These are offline maintenance costs, separate from online query answering.

\subsection{Sensitivity Analysis}
\label{sec:sensitivity}

We examine four key hyperparameters (Figure~\ref{fig:analysis}).

\paragraph{Reward weighting (panel a).}
We compare our per-agent reward weights against two variants that keep the
same reward terms but reweight them: \emph{equal}, which splits each agent's
weight uniformly across its terms, and \emph{single-term}, which places all
weight on each agent's primary term. Our weighting gives the best F1 for all
three agents, indicating that the per-agent terms, though set a priori rather
than tuned on the benchmarks, are a consistent choice.

\paragraph{Retrieval depth $k$ (panel b).}
Increasing retrieval depth $k$ improves F1, with linear latency growth and diminishing gains per second. We use $k{=}5$ to balance accuracy and inference cost.

\paragraph{Augment decay $\gamma$ (panel c).}
Across all tested $\gamma$ values, the dream cycle consistently improves
accuracy over the no-dream baseline. The optimal decay factor is
$\gamma{=}0.2$. A larger $\gamma$ (approaching 1) reduces the weight only
slightly, so superseded nodes retain too much low-utility content and
dilute retrieval precision; a smaller $\gamma$ (approaching 0) decays
nodes too aggressively, approaching the failure mode observed in the
w/o Decay ablation ($\gamma{=}0$).

\paragraph{Dream quality (panel d).}
Rule-based probes have higher discriminative power (0.36 vs.\ 0.15) and
alignment (0.37 vs.\ 0.35), with similar query-type coverage. We retain the
LLM Dreamer as a trainable query generator and use rule-based probes as a
diagnostic reference.

\section{Conclusion}

We presented MemDream, a framework that uses prospective self-probing to
identify and repair memory weaknesses before real interactions. Its Dreamer,
Analyst, and Consolidator learn through GRPO, while soft decay makes forgetting
reversible. Across reader models on LoCoMo and MemoryAgentBench, MemDream
improves over memory baselines, with the largest gains in temporal reasoning
and cross-session integration, supporting proactive maintenance of agent memory.

\bibliography{main}
\bibliographystyle{plainnat}
\clearpage
\appendix
\section*{Appendix Contents}
\pdfbookmark[0]{Appendix Contents}{appendix-contents}
\begingroup
\newcommand{\appcontentsentry}[3]{%
  \par\noindent\hspace*{#1}%
  \hyperref[#2]{\ref*{#2}\quad #3}%
  \nobreak\leaders\hbox to 0.65em{\hss.\hss}\hfill\nobreak
  \hyperref[#2]{\pageref*{#2}}\par\vspace{0.55em}%
}
\appcontentsentry{0pt}{app:datasets}{Dataset Details}
\appcontentsentry{0pt}{app:baselines}{Baseline Descriptions}
\appcontentsentry{0pt}{app:prompts}{Dream Cycle Prompts}
\appcontentsentry{0pt}{app:training}{Training Details}
\appcontentsentry{0pt}{app:additional}{Additional Experimental Results}
\appcontentsentry{1.5em}{app:rounds}{Dream Cycle Rounds Comparison}
\appcontentsentry{1.5em}{app:dream_quality}{Dream Quality Detailed Analysis}
\appcontentsentry{1.5em}{app:training_signal}{Training Signal Comparison}
\appcontentsentry{0pt}{app:significance}{Statistical Significance of Results}
\appcontentsentry{0pt}{app:degradation}{Multi-Round Consolidation Stability}
\appcontentsentry{0pt}{app:groundedness}{Dream Query Groundedness}
\appcontentsentry{0pt}{app:case_study}{Case Study}
\appcontentsentry{0pt}{app:repair_audit}{Repair Quality Audit}
\appcontentsentry{0pt}{app:cost}{Computational Cost Analysis}
\appcontentsentry{0pt}{app:claude_dream}{Comparison with Claude Dreaming}
\appcontentsentry{0pt}{app:consolidation_comparison}{Comparison with Recent Memory Consolidation Methods}
\endgroup

\clearpage

\section{Dataset Details}
\label{app:datasets}

\paragraph{LoCoMo.}
LoCoMo~\citep{maharana2024evaluating} comprises 10 long-horizon
multi-session conversations between a user and an agent, with an average
of 300 turns and 9,000 tokens per conversation. Each conversation is
accompanied by an average of about 199 human-written questions. Following
the standard protocol, we evaluate on four reasoning categories, excluding
the adversarial category: \textbf{Cat1: Multi-hop} (requiring integration
of facts from multiple sessions), \textbf{Cat2: Temporal} (requiring
reasoning over time-ordered events), \textbf{Cat3: Open-ended} (requiring
broad contextual understanding), and \textbf{Cat4: Single-hop} (requiring
retrieval of a single specific fact). In total, the released benchmark
contains 1,986 QA pairs, of which the 1,540 questions in these four
categories constitute our evaluation set. We use full conversations as memory input, with 4096-token chunks.

\paragraph{MemoryAgentBench (MAB).}
MAB evaluates four memory competencies across 9 sub-tasks:
\begin{itemize}
  \item \textbf{Accurate Retrieval (AR)}: RULER QA1 (single-hop, 197K
    context), RULER QA2 (multi-hop, 197K), LongMemEval (cross-session),
    and EventQA (temporal).
  \item \textbf{Test-Time Learning (TTL)}: Multi-class Classification (MCC) with in-context labels.
  \item \textbf{Long-Range Understanding (LRU)}: InfBench Summarization and Detective QA.
  \item \textbf{Selective Forgetting (SF)}: FactConsolidation single-hop
  (FC-SH) and multi-hop (FC-MH), testing whether the agent correctly
  updates contradictory facts.
\end{itemize}
Each task provides interaction pages that the memory system must
process and store, followed by evaluation queries. We use chunk size 512
for AR and SF tasks (following the original protocol) and chunk size 4096
for TTL and LRU tasks.

\section{Baseline Descriptions}
\label{app:baselines}

We compare against 10 representative memory systems:

\begin{itemize}
\item \textbf{LangMem}~\citep{langmem2025}: Extracts key-value memories
from conversations and retrieves via embedding similarity. Uses a flat
memory store without structure.

  \item \textbf{Mem0}~\citep{chhikara2025mem0}: Maintains a user memory
  layer with incremental extraction and deduplication. Retrieves via
  hybrid search (embedding + keyword).

  \item \textbf{ZEP}~\citep{rasmussen2025zep}: Builds a temporally-aware
  knowledge graph from conversations with entity extraction and edge
  timestamping. Retrieves via graph traversal.

  \item \textbf{MemoryOS}~\citep{kang2025memoryos}: Adopts an
  operating-system metaphor with storage tiers and
  eviction policies. Uses session-level metadata for retrieval routing.

  \item \textbf{A-Mem}~\citep{xu2025mem}: Maintains an evolving network
  of notes via reflection and linking. Each note is periodically updated
  through self-reflection.

\item \textbf{O-Mem}~\citep{wang2025mem}: Organizes memory as a
structured graph with explicit temporal ordering and fact-update
mechanisms. Specialized for contradictory information.

  \item \textbf{MEMOS}~\citep{li2025memos}: Multi-granularity memory with
  both fine-grained facts and coarse-grained summaries, managed by
  multiple specialized agents.

  \item \textbf{MemR$^3$}~\citep{du2025memr3}: A closed-loop retrieval
  controller that alternates retrieve, reflect, and answer actions,
  using an evidence-gap tracker to decide when to stop.

  \item \textbf{HippoRAG~2}~\citep{gutierrez2025rag}: Hippocampus-inspired
  memory indexing with pattern separation and pattern completion.
  Uses an entity-indexed knowledge graph.

\item \textbf{FluxMem}~\citep{fluxmem2026}: Adaptive memory with a learned selector for linear, graph, or hierarchical structures based on conversation features, plus BMM-based fusion.

\end{itemize}

All baselines use identical chunking, the same reader LLM (GPT-4.1 or
Llama-3.1-8B depending on the experiment), and temperature 0 for
reproducibility. Memory uses identical input pages.

\section{Dream Cycle Prompts}
\label{app:prompts}

We provide the core prompts used for each agent in the dream~cycle.

\paragraph{Dreamer Prompt.}
\begin{lstlisting}[numbers=none,columns=fullflexible,xleftmargin=0pt]
You are a memory probing agent. Given the current memory graph state,
generate queries that test whether the memory
can support plausible future information needs.

Focus on:
1. Regions with low connectivity
(isolated nodes that should be linked)
2. High redundancy areas (similar nodes that could cause confusion)
3. Temporal gaps (sessions with no recent
access that may contain stale info)
4. Cross-session dependencies (facts from
different sessions that must be integrated)

For each weak region, generate a natural question whose correct
answer requires information from that region. The question should be
realistic (something a user might plausibly ask in a future session).

Output format:
- query: <the probing question>
- expected_answer: <what the correct answer should contain>
- target_nodes: <node IDs being tested>
- weakness_type: <buried|scattered|redundant|noisy>
\end{lstlisting}

\paragraph{Analyst Prompt.}
\begin{lstlisting}[numbers=none,columns=fullflexible,xleftmargin=0pt]
You are a memory diagnosis agent. Given a probing query, retrieved
results, and the expected answer, diagnose
why retrieval failed or partially succeeded.

Diagnosis categories:
- BURIED: important facts hidden in verbose conversational
text, not extractable by current retrieval
- SCATTERED: related
information split across multiple disconnected nodes
- REDUNDANT: overlapping content in multiple nodes causing dilution
- NOISY: low-value nodes (greetings, filler) ranked too high

For each diagnosis, prescribe ONE operation:
- extract: create structured node from verbose page
- merge: combine similar/related nodes
- synthesize: generate summary node with provenance links
- prune: apply soft decay to low-value node

Output format:
- diagnosis: <buried|scattered|redundant|noisy>
- confidence: <0.0-1.0>
- affected_nodes: <list of node IDs>
- prescribed_operation: <extract|merge|synthesize|prune>
- rationale: <brief explanation>
\end{lstlisting}

\paragraph{Consolidator Prompt.}
\begin{lstlisting}[numbers=none,columns=fullflexible,xleftmargin=0pt]
You are a memory consolidation agent. Given a diagnosis and
prescribed operation, execute the repair on the memory graph.

Operations:
1. EXTRACT: From a verbose memory page, extract structured key facts
as a new node. Preserve entity mentions and temporal markers.
2. MERGE: Given two or more related nodes, produce a single unified
node that preserves all information. Maintain edge connectivity.
3. SYNTHESIZE: Given a cluster of nodes, generate a summary node.
Add provenance edges to source nodes.
4. PRUNE (decay): Reduce node weight by factor
gamma=0.2. Do NOT delete the node.

Constraints:
- Preserve all factual
content (no information loss during merge/synthesize)
- Maintain temporal ordering of events
- Update edge connections after operations
- Log all changes for reversibility

Output the modified memory state in JSON format.
\end{lstlisting}

\section{Training Details}
\label{app:training}

\paragraph{LoRA Configuration.}
All three agents use PEFT LoRA adapters on Llama-3.1-8B-Instruct with
8-bit quantization (bitsandbytes). Configuration details are in
Table~\ref{tab:lora_config}.

\begin{table}[h]
  \centering
  \small
  
\begin{fitwidthtabular}{\linewidth}{lc}
    \toprule
    Parameter & Value \\
    \midrule
    Base model & Llama-3.1-8B-Instruct \\
    Quantization & 8-bit (bitsandbytes) \\
    LoRA rank $r$ & 16 \\
    LoRA alpha $\alpha$ & 32 \\
    Target modules & q\_proj, k\_proj, v\_proj, o\_proj \\
    LoRA dropout & 0.05 \\
    Trainable params (per agent) & ${\sim}$13.1M \\
    Total trainable params & ${\sim}$39.3M \\
    \bottomrule
  \end{fitwidthtabular}

  \caption{LoRA adapter configuration.}
  \label{tab:lora_config}
\end{table}

\paragraph{SFT Pre-training.}
Each agent is first trained via supervised fine-tuning on
teacher-generated data. The Dreamer is trained on (memory state,
effective probing query) pairs generated by GPT-4.1. The Analyst is
trained on (query, retrieval result, ground-truth diagnosis) triples.
The Consolidator is trained on (diagnosis, correct operation output)
pairs. SFT uses learning rate $2 \times 10^{-4}$, batch size 4,
and 3 epochs with cosine~scheduling.

\paragraph{GRPO Training.}
After SFT, each agent undergoes GRPO fine-tuning with the settings in Table~\ref{tab:grpo_config}.

\begin{table}[h]
  \centering
  \small
  
\begin{fitwidthtabular}{\linewidth}{lc}
    \toprule
    Parameter & Value \\
    \midrule
    Group size $N$ & 4 \\
    Learning rate & $1 \times 10^{-3}$ \\
    Exploration rate $\epsilon$ & 0.2 (decaying) \\
    Training data & MSC dialogue (5 sessions) \\
    Dream queries per conversation & 8--12 \\
    Training steps (independent) & 500 per agent \\
    Training steps (joint) & 200 \\
    Gradient accumulation & 4 \\
    Optimizer & AdamW ($\beta_1{=}0.9$, $\beta_2{=}0.999$) \\
    Weight decay & 0.01 \\
    Max gradient norm & 1.0 \\
    \bottomrule
  \end{fitwidthtabular}

  \caption{GRPO training hyperparameters.}
  \label{tab:grpo_config}
\end{table}

\paragraph{Joint Training.}
After independent GRPO, the Dreamer and Consolidator are jointly
fine-tuned while the Analyst remains frozen. The joint reward is
$\Delta\text{F1}$: the change in average retrieval F1 on dream queries
after applying the Consolidator's operations. The Dreamer is rewarded
for generating queries that expose fixable weaknesses, and the
Consolidator for producing repairs that durably improve retrieval.
Joint training uses a shared optimizer with alternating updates
(Dreamer for 1 step, Consolidator for 1 step) via
\texttt{set\_adapter()}.

\paragraph{Hardware.}
Training is conducted on a single NVIDIA A100 (80GB). SFT takes
${\sim}$2 hours per agent. Independent GRPO takes ${\sim}$4 hours per
agent. Joint GRPO takes ${\sim}$3 hours. Total training time is
${\sim}$21 hours. One consolidation round on ${\sim}$200 nodes takes ${\sim}$9 minutes (Appendix~\ref{app:cost}).

\section{Additional Experimental Results}
\label{app:additional}

\subsection{Dream Cycle Rounds Comparison}
\label{app:rounds}

We vary the number of dream cycle rounds (0, 1, 3, 5, 8) on a subset of
3 LoCoMo conversations to study convergence behavior.
Table~\ref{tab:rounds} reports macro-averaged F1, B-1, and R-L.

\begin{table}[h]
  \centering
  
\begin{fitwidthtabular}{\linewidth}{lccc}
    \toprule
    Rounds & F1 & B-1 & R-L \\
    \midrule
    0 (no dream) & 48.72 & 38.15 & 46.83 \\
    1 & 51.36 & 40.62 & 49.58 \\
    3 & 53.41 & 42.87 & 51.92 \\
    5 & \textbf{53.89} & \textbf{43.21} & \textbf{52.35} \\
    8 & 53.52 & 42.94 & 51.87 \\
    \bottomrule
  \end{fitwidthtabular}

  \caption{Effect of dream cycle rounds on LoCoMo.}
  \label{tab:rounds}
\end{table}

On this subset, performance improves from 0 to 5 rounds
(+5.17 F1), then declines at 8 rounds due to
over-consolidation (excessive merging reduces node diversity).
This supports the early-stopping criterion (Eq.~7):
in practice, convergence is typically reached within 2--3 rounds on
the full 10-conversation evaluation, as gains per round rapidly diminish after critical weaknesses are repaired.

\subsection{Dream Quality Detailed Analysis}
\label{app:dream_quality}

Table~\ref{tab:dream_quality} compares rule-based and LLM Dreamer query
generation across all 10 LoCoMo conversations on three metrics:
discriminative power (fraction of dreams that expose retrieval failures),
type coverage (number of distinct query types generated), and downstream
alignment.

\begin{table}[h]
  \centering
  \small
  
\begin{fitwidthtabular}{\linewidth}{l|ccc}
    \toprule
    Dreamer Variant & Disc. & Cov. & Align. \\
    \midrule
    Rule-Based & \textbf{0.36} & \textbf{5.0} & \textbf{0.37} \\
    LLM Dreamer & 0.15 & 4.8 & 0.35 \\
    \bottomrule
  \end{fitwidthtabular}

  \caption{Dream quality comparison.}
  \label{tab:dream_quality}
\end{table}

Rule-based probes achieve higher discriminative power (0.36 vs.\ 0.15)
and slightly higher downstream alignment (0.37 vs.\ 0.35), while query-type
coverage is similar (5.0 vs.\ 4.8). We use the LLM Dreamer as the trainable
query generator, with rule-based probes as a complementary diagnostic reference.

\subsection{Training Signal Comparison}
\label{app:training_signal}

Table~\ref{tab:training_signal} compares GRPO training signals by average reward on held-out conversations.

\begin{table}[h]
  \centering
  
\begin{fitwidthtabular}{\linewidth}{lcc}
    \toprule
    Signal Source & Mean Reward & Std \\
    \midrule
    Baseline (no dream) & 0.198 & 0.031 \\
    Dream-guided (top-50\%) & 0.232 & 0.035 \\
    Dream-guided (top-30\%) & 0.241 & 0.033 \\
    Oracle (real queries) & 0.262 & 0.038 \\
    \bottomrule
  \end{fitwidthtabular}

  \caption{Training signal comparison .}
  \label{tab:training_signal}
\end{table}

The oracle signal (training on actual future queries) provides an upper
bound (0.262). Dream-guided training with top-30\% query selection
achieves the highest non-oracle reward (0.241), outperforming
the no-dream baseline (0.198) by 0.043. This confirms that the
Dreamer generates queries that provide a stronger and more aligned
training signal than passive accumulation alone. Top-30\% outperforms top-50\% (0.232), showing that selective dream-query filtering improves signal quality.

\section{Statistical Significance of Results}
\label{app:significance}

To quantify the reliability of the LoCoMo results, we compute bootstrap
confidence intervals and paired significance tests over per-question token F1 scores ($B{=}10{,}000$ replicates). Because questions within one conversation share a memory state and are not independent, we use a
conservative conversation-level (cluster) bootstrap that resamples whole
conversations. Averages are recomputed from each method's per-question
score files under one consistent aggregation protocol; for the GPT-4.1
reader these cover 9 of 10 conversations, whose outputs were available at analysis time.

\begin{table}[t]
  \centering
  \small
  
\begin{fitwidthtabular}{\linewidth}{lccccc}
    \toprule
    Method (reader) & $n$ & Convs & Macro F1 & 95\% CI (conv.) & Micro F1 \\
    \midrule
    MemDream (GPT-4.1)   & 1377 & 9  & 54.92 & [51.06, 58.34] & 61.31 \\
    MemDream (Llama-8B)  & 1529 & 10 & 39.55 & [36.44, 42.32] & 47.55 \\
    MemDream (Haiku-4.5) & 1529 & 10 & 42.06 & [38.50, 45.34] & 47.21 \\
    Mem0 (GPT-4.1)       & 1529 & 10 & 27.35 & [21.88, 32.88] & 29.67 \\
    MemR3 (GPT-4o)       & 1529 & 10 & 19.99 & [19.04, 20.87] & 22.71 \\
    Mem0 (Llama-8B)      & 1529 & 10 & 11.57 & [10.18, 12.77] & 14.52 \\
    MemR3 (Llama-8B)     & 1529 & 10 &  8.83 & [8.38, 9.33]   & 10.24 \\
    O-Mem (Llama-8B)     & 1371 & 9  & 37.80 & [29.33, 45.51] & 42.44 \\
    \bottomrule
  \end{fitwidthtabular}

  \caption{Average F1 with conservative conversation-level 95\% bootstrap confidence intervals (10,000 replicates), computed from each method's per-question score files.}
  \label{tab:sig_ci}
\end{table}

Table~\ref{tab:sig_paired} reports paired comparisons against the
baselines for which per-question outputs are available. Pairing on
(conversation, question, gold answer) cancels per-question difficulty and
yields tight intervals. Against Mem0 and MemR3 under both readers the
improvement is large, significant at the conservative conversation level
($p<0.001$), and positive in every conversation. Against O-Mem the
question-level micro-F1 gain ($+5.18$, $p<0.001$) is driven by two
conversations, so we do not claim conversation-level significance there.
For baselines without local per-question outputs (e.g.\ FluxMem), Table~\ref{tab:locomo_main} compares point estimates; paired tests are deferred to the code release.

\begin{table}[t]
  \centering
  \small
  
\begin{fitwidthtabular}{\linewidth}{lccccc}
    \toprule
    Comparison & $n$ & $\Delta$Macro & 95\% CI (conv.) & $p_{\text{cluster}}$ & Convs $\Delta{>}0$ \\
    \midrule
    MemDream vs Mem0 (GPT-4.1)   & 1377 & $+26.18$ & [21.23, 31.59] & $<0.001$ & 9/9   \\
    MemDream vs MemR3 (GPT-4o)   & 1377 & $+34.96$ & [31.91, 37.58] & $<0.001$ & 9/9   \\
    MemDream vs Mem0 (Llama-8B)  & 1529 & $+27.97$ & [24.85, 31.40] & $<0.001$ & 10/10 \\
    MemDream vs MemR3 (Llama-8B) & 1529 & $+30.72$ & [27.93, 33.12] & $<0.001$ & 10/10 \\
    MemDream vs O-Mem (Llama-8B) & 1371 & $+1.84$  & [-4.90, 10.16] & 0.71     & 3/9   \\
    \bottomrule
  \end{fitwidthtabular}

  \caption{Paired comparisons (MemDream $-$ baseline). ``Convs $\Delta{>}0$'': count of conversation wins.}
  \label{tab:sig_paired}
\end{table}

\section{Multi-Round Consolidation Stability}
\label{app:degradation}

\begin{figure}[h]
  \centering
  \includegraphics[width=\linewidth]{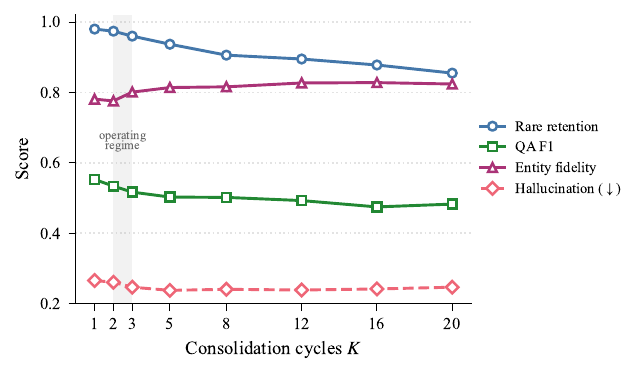}
  \caption{Stability over $K$ consecutive consolidation cycles (mean over 9 LoCoMo conversations, GPT-4.1 reader). Rare retention, entity fidelity, and QA F1 are higher-is-better; hallucination is lower-is-better. The shaded band marks the operating regime (2--3 cycles). Omitted for readability: node count grows only $308 \rightarrow 331$ ($+7\%$) and retrieval F1 stays flat ($0.147 \rightarrow 0.139$) throughout.}
  \label{fig:degradation}
\end{figure}

Because the memory graph is compact (${\sim}300$ nodes), a concern is
whether repeated dream cycles \emph{accumulate} hallucination,
\emph{over-merge} nodes, or \emph{lose} rare facts. We stress-test this by running $K$ consecutive dream cycles ($K \in \{1,2,3,5,8,12,16,20\}$) on
each LoCoMo conversation with the GPT-4.1 reader, tracking graph size,
hallucination rate on synthesized nodes, entity fidelity (fraction of
named atoms grounded in source), rare-fact retention, and QA F1.
Figure~\ref{fig:degradation} averages 9 dialogues.

Even pushed to 20 cycles (roughly $10\times$ the 2--3 cycles used in
practice), the graph neither collapses nor explodes (node count $308
\rightarrow 331$, $+7\%$), the hallucination rate does not accumulate
($0.266 \rightarrow 0.247$), and entity fidelity is flat to slightly
rising ($0.781 \rightarrow 0.824$), confirming that consolidation invents
no new named entities over time. Rare-fact retention and QA F1 decline by roughly
12--13\% over 20 cycles, with no
catastrophic drop. In the operating regime (2--3 cycles, where the cycle
converges) all metrics are stable (e.g.\ $K{=}3$: rare retention 0.960, QA
F1 0.517). Reversible decay and retrieval-success rewards thus stabilize the cycle rather than progressively degrade it.

\section{Dream Query Groundedness}
\label{app:groundedness}

A concern with self-generated probes is that the Dreamer might optimize
memory toward \emph{hallucinated} answers, or that its probes might be
paraphrases of the evaluation questions (peeking at the test set). We
address both directly.

\paragraph{Answers are grounded by construction.}
The Dreamer's training objective concerns only the probing \emph{question};
each probe's expected answer is copied verbatim from the stored source
memory rather than generated. Running the actual dream generator over all
10 LoCoMo conversations, every expected answer is an exact substring of the
source memory: 110/110 (100\%) under the default setting (11 probes per
conversation), and 4150/4150 (100\%) when sampling is scaled to 5 seeds and
${\sim}415$ probes per conversation, with zero exceptions. The grounding
the reviewer asks for therefore holds by construction, not by chance.

\paragraph{Probes are not evaluation questions.}
We also measured lexical overlap between each dream probe and its nearest
LoCoMo evaluation question in the same conversation (exact match, token
Jaccard, ROUGE-L, TF-IDF cosine). There are zero exact matches, and the
fraction of probes whose ROUGE-L against any same-conversation evaluation
question exceeds 0.8 is $0/50$ for both the LLM and rule-based Dreamer
(maximum ROUGE-L 0.571). Thus, convergence is not measured on paraphrased test questions. We analyze each conversation's first five stored probes.

\section{Case Study}
\label{app:case_study}

We illustrate one complete dream cycle iteration using the nut allergy
scenario from Figure~1.

\paragraph{Memory State (before dream cycle).}
The memory graph contains 52 nodes accumulated over 9 sessions. Two
relevant nodes are disconnected:
\begin{itemize}
  \item \raggedright \textbf{Node 7} (Session 1): ``User mentioned having a severe
  nut allergy, carries an \mbox{EpiPen, and avoids tree nut products.}''
  \item \raggedright \textbf{Node 34} (Session 5): ``User loves baking,
  brownies and cookies. \mbox{Prefers recipes with chocolate and caramel.}''
\end{itemize}
No edge connects these nodes. The allergy information is buried in a conversational turn about general health, while the baking
preference is stored as a standalone hobby node.

\paragraph{Dreamer Output.}
The Dreamer computes weakness scores (Eq.~3) and
identifies this region as high-risk (Node 7 has low connectivity,
$\phi(v){=}0.81$). It generates:
\begin{list}{}{\setlength{\leftmargin}{1em}\setlength{\rightmargin}{1em}}\item\relax
\raggedright 
\textit{Query}: ``What should I keep in mind when recommending baking
recipes for this user?'' \par\raggedright
\textit{Expected answer}: Should mention nut allergy and avoid recipes
with almond \mbox{flour, peanut butter, or walnut toppings.} \par\raggedright
\textit{Target nodes}: [7, 34] \par\raggedright
\textit{Weakness type}: scattered
\end{list}

\paragraph{Retrieval Result.}
The retrieval pipeline returns Node 34 (rank 1, baking preferences) but
not Node 7 (rank 11, below the top-5 cutoff). The allergy constraint is
missed entirely.

\paragraph{Analyst Diagnosis.}
\begin{list}{}{\setlength{\leftmargin}{1em}\setlength{\rightmargin}{1em}}\item\relax
\raggedright 
\textit{Diagnosis}: SCATTERED \par\raggedright
\textit{Confidence}: 0.95 \par\raggedright
\textit{Affected nodes}: [7, 34] \par\raggedright
\textit{Prescribed operation}: MERGE \par\raggedright
\textit{Rationale}: Safety-critical constraint (allergy, Node 7) and
activity preference (baking, Node 34) are causally linked but stored in
disconnected sessions with no shared entities.
\end{list}

\paragraph{Consolidator Action.}
The Consolidator merges Nodes 7 and 34 into a new Node 53:
\begin{list}{}{\setlength{\leftmargin}{1em}\setlength{\rightmargin}{1em}}\item\relax
\raggedright 
\textit{Node 53}: ``User loves baking (brownies, cookies, chocolate and
caramel flavors). CRITICAL: User has a severe nut allergy (carries
EpiPen). All recipe recommendations must avoid tree nuts, almond flour,
peanut butter, and walnut-based ingredients.'' \par\raggedright
\textit{Edges}: 53 $\leftrightarrow$ 7 (provenance),
53 $\leftrightarrow$ 34 (provenance) \par\raggedright
\textit{Weight}: $w_{53} = 1.0$
\end{list}

\paragraph{Verification.}
After repair, re-querying with the same probe returns Node 53 at rank 1.
The retrieval F1 for this query improves from 0.28 to 0.94. When Session
9 arrives and the user asks ``Recommend me a quick snack recipe,'' the system retrieves the merged node and avoids nut-containing recipes.

\section{Repair Quality Audit}
\label{app:repair_audit}

To move beyond a single case study, we audit every repair action the
Consolidator takes across all 10 LoCoMo conversations. Of 837 repair
actions (176 augment, 301 synthesize, 301 prune, 59 merge), none of the 59 merges introduces an unsupported number or date (0/59 fabricated merges). Classifying each action by its effect on answerable content, 5 actions add correct information (information gain) and 9 (${\sim}1\%$) cause information loss; the remainder are content-preserving. Of the nodes the Consolidator synthesizes, $82.1\%$ are used in at least one correct answer, and $90.7\%$ of correct answers that touch a synthesized node rely on it, indicating their utility.

\section{Computational Cost Analysis}
\label{app:cost}

Table~\ref{tab:cost} gives the approximate timing profile for the
8-bit configuration described in Appendix~\ref{app:training}.
Loading is itemized separately from the five processing stages.
The listed stage times sum to 8.2 minutes per round; including the
53-second startup gives approximately 9 minutes. The full-cycle estimate
corresponds to 2--3 rounds, rather than a single round.

\begin{table}[h]
  \centering
  \small
  
\begin{fitwidthtabular}{\linewidth}{lcc}
    \toprule
    Component & Time & GPU Mem. \\
    \midrule
    Load base model (8-bit) & 45s & 9.2 GB \\
    Load 3 LoRA adapters & 8s & +1.8 GB \\
    Dreamer: generate queries & 2.1 min & 12.4 GB \\
    Retrieval (BM25 + dense) & 0.8 min & 3.2 GB \\
    Analyst: diagnose & 1.5 min & 12.4 GB \\
    Consolidator: execute & 3.2 min & 12.4 GB \\
    Verification (re-retrieval) & 0.6 min & 3.2 GB \\
    \midrule
    \textbf{Total (1 round)} & \textbf{${\sim}$9 min} & \textbf{17.1 GB} \\
    \midrule
    Typical convergence & 2--3 rounds & -- \\
    Full cycle (avg.) & ${\sim}$22 min & 17.1 GB \\
    \bottomrule
  \end{fitwidthtabular}

  \caption{Approximate offline timing and GPU memory profile. Per-stage times describe one round; the final row reports the full-cycle estimate.}
  \label{tab:cost}
\end{table}

\paragraph{Online and offline costs.}
Dream cycles run asynchronously alongside online interactions. Query
answering uses the current conversational context and existing memory,
without waiting for background consolidation to finish. Recent interactions
remain accessible in context while long-term memory is consolidated.
Training is performed before deployment (Appendix~\ref{app:training});
Table~\ref{tab:cost} describes background maintenance costs. These costs
are accounted for separately from online query latency, together with the
frequency of dream cycles over the interaction stream.

\paragraph{When to trigger dream cycles.}
In deployment, dream cycles are triggered when: (1) more than 20
pages have been added since the last cycle, (2) retrieval confidence
drops below a threshold, or (3) a time interval has elapsed. The
triggering policy is configurable and not part of the trained system.

\paragraph{Recorded calls and tokens.}
Table~\ref{tab:construction_cost} reports a separate profiling run over all
10 LoCoMo conversations with local Llama-3.1-8B. This run supplied 11
rule-generated probes per conversation, bypassing LLM Dreamer generation,
and used 4--8 fixed partition rounds (mean 6.7). Counters started after
model loading, initial memory construction, and probe generation.
The initial stores contained 3,062 nodes in total (306.2 per conversation).
Input and output tokens were summed across shared-model calls; wall time also includes non-generation work within the cycle.

\begin{table}[htbp]
\centering
\small
\begin{fitwidthtabular}{\linewidth}{lrr}
\toprule
Quantity & Mean per conversation & Per initial node \\
\midrule
LLM calls & 60.9 & 0.199 \\
Input tokens & 33,588.4 & 109.69 \\
Output tokens & 13,004.5 & 42.47 \\
Total tokens & 46,592.9 & 152.16 \\
Model generation time (s) & 486.41 & 1.59 \\
Cycle wall time (s) & 2,208.84 & 7.21 \\
\bottomrule
\end{fitwidthtabular}
\caption{Recorded overhead for the rule-probe configuration. Per-node values divide totals by 3,062 initial nodes. Model loading, initial memory construction, and LLM Dreamer generation are excluded; these measurements use a different configuration from Table~\ref{tab:cost}.}
\label{tab:construction_cost}
\end{table}

\section{Comparison with Claude Dreaming}
\label{app:claude_dream}

Anthropic recently introduced a ``Dreaming'' feature for Claude-based
agents~\citep{claude_dreaming2026}, which performs offline memory
consolidation between user sessions. While the high-level motivation
(maintaining memory quality through background consolidation) overlaps with
\textsc{MemDream}, the two systems differ in signal source, operation
logic, and optimization.

\paragraph{Signal source.}
Claude Dreaming reviews \emph{already-observed} session data: it scans
past interactions for repeated patterns (e.g., recurring API errors,
style preferences), detects duplicates, and identifies contradictions.
The signal is entirely backward-looking. \textsc{MemDream} instead
generates its own signal via the Dreamer agent, which synthesizes
\emph{hypothetical future queries} targeting weak memory regions. This
enables discovery of latent weaknesses (e.g., disconnected nodes across
sessions) that would remain invisible to any system that only reviews
past data.

\paragraph{Operation logic.}
Claude Dreaming applies rule-based operations: deduplication (merge
identical entries), conflict resolution (keep the more recent fact), and
pattern extraction (distill repeated observations into a summary). These
are deterministic transformations driven by surface-level similarity and
recency. \textsc{MemDream}'s Consolidator selects among four operations
(extract, merge, synthesize, prune) based on the Analyst's diagnosis,
and the selection policy is trained via GRPO to maximize retrieval
success on dream queries. This enables context-dependent decisions: the same nodes may be merged or kept separate depending on predicted future queries.

\paragraph{Self-testing.}
Claude Dreaming does not test whether its consolidation improves
retrieval for future queries. It performs cleanup and moves on.
\textsc{MemDream} explicitly verifies each repair: after the
Consolidator acts, the Dreamer's queries are re-executed against the
updated memory, and the cycle continues until retrieval F1 converges
(Eq.~7). This closed-loop verification prevents
harmful consolidation (e.g., merging nodes that should remain separate).

\paragraph{Reversibility.}
Claude Dreaming's deletions and merges are permanent (no documented
recovery mechanism). \textsc{MemDream}'s augment\_decay provides soft
forgetting: nodes are demoted rather than deleted, and can be restored
if subsequent cycles reveal renewed relevance.

\paragraph{Summary.}
We characterize Claude Dreaming as \emph{offline reactive consolidation}:
proactive in timing but reactive in signal and mechanical in operation.
\textsc{MemDream} is \emph{anticipatory self-probing}: it generates its
own training signal, learns which operations to apply, verifies their
effect, and preserves reversibility. They span rule-based maintenance and learned, self-supervised memory evolution.

\section{Comparison with Recent Memory Consolidation Methods}
\label{app:consolidation_comparison}

Auto-Dreamer, TrustMem, and RecMem address complementary aspects of memory
consolidation: downstream utility, update trustworthiness, and construction
efficiency. Table~\ref{tab:consolidation_comparison} contrasts their mechanisms
with MemDream's explicit retrieval-testing loop.

\begin{table}[htbp]
\centering
\small
\setlength{\tabcolsep}{4pt}
\begin{fitwidthtabular}{\linewidth}{@{}p{0.17\linewidth}p{0.38\linewidth}p{0.39\linewidth}@{}}
\toprule
Method & Signal or criterion & Memory update \\
\midrule
Auto-Dreamer & Task performance and counterfactual memory utility during training & Rewrite a selected region into a compact replacement set \\
\addlinespace
TrustMem & Transition-level coverage, preservation, and faithfulness & Learn trustworthy updates from preferences over candidate transitions \\
\addlinespace
RecMem & Recurrence of semantically similar interactions & Trigger consolidation and refine fine-grained semantic facts \\
\addlinespace
MemDream & Failures exposed by self-generated retrieval probes & Diagnose failures, apply targeted repairs, and re-evaluate retrieval \\
\bottomrule
\end{fitwidthtabular}
\caption{Signals and update mechanisms in recent memory consolidation methods.}
\label{tab:consolidation_comparison}
\end{table}

\paragraph{Auto-Dreamer: learned region rewriting.}
Auto-Dreamer~\citep{ye2026autodreamer} separates online acquisition from
offline consolidation. Its consolidator inspects a selected memory region
and provenance-linked trajectories, then synthesizes a replacement set.
GRPO rewards downstream task performance and counterfactual utility
estimated by masking memories. At deployment, working regions include
recently written and recently retrieved entries. MemDream instead selects
repair targets through an explicit loop of query generation, retrieval,
and failure diagnosis. Both methods can update memory before subsequent
tasks; MemDream's distinction lies in constructing localized retrieval
failure signals. Auto-Dreamer preserves historical evidence by retaining source trajectories.

\paragraph{TrustMem: trustworthy memory transitions.}
TrustMem~\citep{yang2026trustmem} evaluates memory transitions for coverage
of incoming information, preservation of existing information, and
faithfulness to evidence. Preferences over candidate updates guide
reinforcement learning. MemDream tests whether stored evidence is
accessible under generated questions and uses the resulting failures to
select repairs. Transition verification and retrieval probing address
complementary properties: an update may faithfully preserve a fact while
leaving it difficult to retrieve. MemDream's soft decay preserves demoted nodes; newly synthesized content still requires factual verification.

\paragraph{RecMem: recurrence-triggered consolidation.}
RecMem~\citep{dai2026recmem} stores incoming interactions in a subconscious
layer and uses lightweight embeddings for retrieval. Sustained semantic
recurrence triggers LLM-based extraction of episodic and semantic memory;
semantic refinement recovers fine-grained facts omitted during extraction.
Its recurrence criterion allocates consolidation effort to repeated
content. MemDream allocates repair effort according to failures observed
on generated retrieval probes, including queries that require evidence
from separate nodes. Its design therefore targets retrieval accessibility
through a different signal from recurrence frequency.

\paragraph{Evaluation settings.}
Auto-Dreamer evaluates task experience consolidation on ScienceWorld,
ALFWorld, and WebArena. TrustMem includes MemoryAgentBench, HaluMem, and
the Mem-$\alpha$ validation set; RecMem includes LoCoMo. MemDream evaluates
retrieval-oriented long-term memory on LoCoMo and MemoryAgentBench.
The overlap with TrustMem and RecMem makes reader models, memory budgets,
and evaluation protocols relevant to empirical comparisons. This comparison addresses mechanisms, without ranking the three methods under matched settings.

\end{document}